\documentclass{article}

\usepackage{microtype}
\usepackage{graphicx}
\usepackage{subfigure}
\usepackage{amsmath}
\usepackage{amssymb}
\usepackage{comment}
\usepackage{multirow}
\usepackage{hyperref}
\usepackage{booktabs} 

\usepackage{hyperref}

\usepackage[accepted]{icml2018}

\newcommand{\modelname}{GBP }
\newcommand{\modelnamep}{DistGBP }

\icmltitlerunning{Model-Based Planning with Discrete and Continuous Actions}

\begin{document}

\twocolumn[
\icmltitle{Model-Based Planning with Discrete and Continuous Actions}




\begin{icmlauthorlist}
\icmlauthor{Mikael Henaff}{nyu}
\icmlauthor{Will Whitney}{nyu}
\icmlauthor{Yann LeCun}{nyu,fb}
\end{icmlauthorlist}

\icmlaffiliation{nyu}{New York University}
\icmlaffiliation{fb}{Facebook AI Research}

\icmlcorrespondingauthor{Mikael Henaff}{mbh305@nyu.edu}

\icmlkeywords{Machine Learning, ICML}

\vskip 0.3in
]



\printAffiliationsAndNotice{}  

\begin{abstract}
Action planning using learned and differentiable forward models of the world is a general approach which has a number of desirable properties, including improved sample complexity over model-free RL methods, reuse of learned models across different tasks, and the ability to perform efficient gradient-based optimization in continuous action spaces.
However, this approach does not apply straightforwardly when the action space is discrete.
In this work, we show that it is in fact possible to effectively perform planning via backprop in discrete action spaces, using a simple paramaterization of the actions vectors on the simplex combined with input noise when training the forward model.
Our experiments show that this approach can match or outperform model-free RL and discrete planning methods on gridworld navigation tasks in terms of performance and/or planning time while using limited environment interactions, and can additionally be used to perform model-based control in a challenging new task where the action space combines discrete and continuous actions. We furthermore propose a policy distillation approach which yields a fast policy network which can be used at inference time, removing the need for an iterative planning procedure.

\end{abstract}

\section{Introduction}
\label{submission}

Planning actions in order to accomplish a specific goal is a challenge of fundamental interest in artificial intelligence.
One of the main paradigms for addressing planning problems is model-free reinforcement learning (RL), a general approach where an agent samples actions according to an internal policy and then adjusts the policy as a function of the reward it receives for different actions. This method has the advantage of making minimal assumptions about the task at hand and can learn complex policies using only the raw state representation and a scalar reward signal.
In recent years model-free RL using deep neural network controllers has proven successful for a number of applications including Atari games \citep{mnih15}, robotic manipulation \citep{Gu2016}, navigation and reasoning tasks \citep{Oh2016}, and machine translation \citep{RL_NMT}.
However, it also suffers from several limitations, including high sample complexity \citep{Schulman2015}, unstable training due to a difficult temporal credit assignment problem and non-stationary input distribution, and sensitivity to hyperparameters and implementatation details \citep{DeepRLMatters}.

Model-based planning assumes the existence of a forward model of the environment which can predict how the world will evolve in response to different actions.
Actions can then be planned by using this forward model to select a sequence of actions which will take the agent from its current state to a desired goal state or maximize rewards along a trajectory.
If the actions are discrete, tree search methods can be used to search over different action sequences and evaluate their quality using the forward model.
Expanding a full tree is often computationally infeasible, hence stochastic approximations are often used to only expand its most promising branches \citep{MCTS}.
If actions are continuous, using differentiable forward models is particularly appealing as they provide gradients which define a direction of improvement for a sequence of actions.
These gradients with respect to plans make it possible to do "planning by backprop", directly optimizing a sequence of actions by backpropagating gradients from a goal state through a learned model to update a plan.

Learning a forward model typically has lower sample complexity than model-free RL due to the rich information content of its high-dimensional error signal, and can in some cases be done using observational data. This can provide a way to derive plans in a way that is sample efficient with regard to environment interactions.
When learning policies using a fast simulator, it is possible to try many actions within the simulated environment since they carry little computational cost and mistakes do not affect the real world.
However, when training policies in real environments, minimizing the number of interactions with the environment can often be crucial, as performing actions (such as driving a vehicle or moving a robot) can be orders of magnitude slower than performing an update of the policy model, and mistakes can carry real-world costs.


In this work, we show that planning with discrete and continuous actions using a learned forward model can be done using the same unified gradient-based approach. By using a simple reparameterization of discrete action vectors in the simplex combined with the addition of input noise when training an action-conditional forward model, we obtain a modified loss function in which it is easy to optimize discrete actions by gradient descent. We show experimentally that optimal control methods can then be effectively used in discrete action spaces, and are able to achieve similar performance to a strong tree search baseline while being straightforward to parallelize.
We also show that it is possible to further speed up planning at inference time by training a feedforward policy network to imitate high-quality trajectories generated by gradient descent using the learned forward model and states it was trained on. This can be done using only trajectories synthesized by the forward model and does not require additional environment interaction.
We additionally introduce a challenging task which requires jointly optimizing discrete and continuous actions, and show that our approach is able to learn behaviors which account for complex environment dynamics, outperforming a model-free approach.




\begin{figure}[t!]
\vskip 0.2in
\begin{center}
\includegraphics[width=\columnwidth]{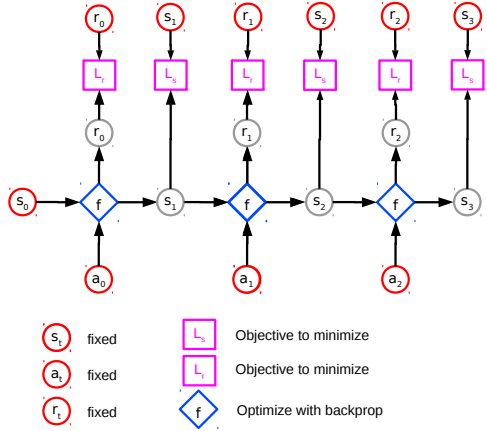}
\caption{Training a forward model. Here $s_t, a_t, r_t$ are taken from the agent's experiences or an observational dataset.}
\label{fwd-train}
\end{center}
\vskip -0.2in
\end{figure}

\section{Planning with Forward Models}

The discrete-time Markov Decision Process (MDP) \citep{Bellman57} is a widely used model for decision making and planning.
In this framework, at every time step $t$ an agent is presented with a representation $s_t$ of the state of the environment, performs an action $a_t$, and receives a reward $r_t$.
The next state is then given as some unknown function of the current state and action, and the process is repeated.

Planning with a forward model requires first estimating both the transition function and the reward function, using the agent's own experience or observational data.
More precisely, the agent learns a transition model $f_S(s_t, a_t)$ which predicts the next state $\tilde{s}_{t+1}$ and a reward model $f_R(s_t, a_t)$ which predicts the next reward $\tilde{r}_{t}$ by minimizing the following loss function over a dataset of trajectories $\mathcal{E}$ indexed by $t$:

\begin{equation*}
\begin{aligned}
& \mathcal{L}(\theta_S, \theta_R) =
& & \frac{1}{M}\sum_{t=1}^M \ell(s_t, a_t, r_t, s_{t+1}) \\
\end{aligned}
\end{equation*}

where the per-sample loss is given by:

\begin{multline}
\ell(s_t, a_t, r_t, s_{t+1}) = \\
\mathcal{L}_S(s_{t+1}, f_S(s_t, a_t)) + \mathcal{L}_R(r_t, f_R(s_t, a_t))
\end{multline}

Here $\mathcal{L}_S$ and $\mathcal{L}_R$ are loss functions which are appropriate to the states and rewards being considered, we use mean-squared error in this work but this choice is dependent on the task. In practice, $f_R$ and $f_S$ may share parameters, for example by sharing the same encoding of the state and action and mapping it to a next state and reward respectively. It can also be beneficial to replace true inputs $s_t$ beyond the initial state by the predicted inputs $\tilde{s}_t$ during training, as shown in Figure \ref{fwd-train}. This has the effect of making the model more robust by training it to account for its future prediction errors. We use this setup in all our experiments.

\begin{figure}[t!]
\vskip 0.2in
\begin{center}
\includegraphics[width=\columnwidth]{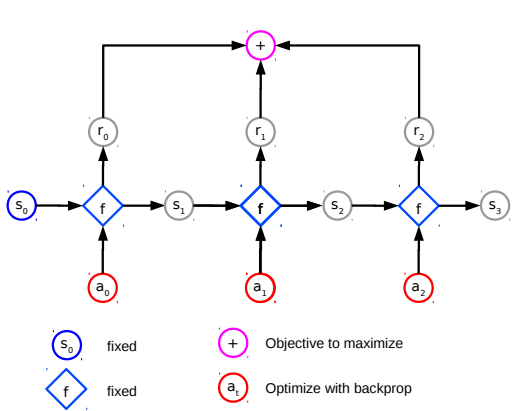}
\caption{Planning actions with a forward model from an initial state $s_0$. Here actions are optimized by gradient descent through the forward model to maximize the sum of rewards.}
\label{fwd-plan}
\end{center}
\vskip -0.2in
\end{figure}


After training, given some current state $s_0$ the agent can use the forward model to plan actions with the goal of maximizing the sum of future rewards.
This is done by solving the following optimization problem:

\begin{equation}
\begin{aligned}
& \underset{a_1, ..., a_T}{\text{arg max}}
& & \sum_{t=1}^T f_R(f_S(\tilde{s}_{t-1}, a_{t-1}), a_t) \\
\end{aligned}
\end{equation}

where $\tilde{s}_0 = s_0$ and $\tilde{s}_{t+1} = f_S(\tilde{s_t}, a_t)$. If $f$ is a neural network model and the input action space is continuous, this optimization procedure can be done using gradient descent by straightforwardly applying the backpropagation rule which computes the gradients with respect to both the weights and the inputs.
Starting from a randomly initialized sequence of action vectors, the loss function can be minimized by performing gradient descent in the input action space.
One can also repeat this process (in sequence or in parallel) for multiple initially sampled sequences of action vectors and pick the one with the highest estimated reward.
This can be useful if the initial action trajectory is of low quality and there is no direction for it to improve, and can be viewed as performing multiple action rollouts, correcting them through gradient descent and keeping the best one.

\section{Planning with Discrete Actions}

When the action space is discrete, input actions are typically encoded as one-hot vectors.
In this case, there is no guarantee that the result of the above optimization procedure will constitute a valid input action sequence.
We propose two modifications to remedy this issue.
The first is to restrict the set of valid actions to the simplex.
Let $\mathcal{A} = \{e_1,...,e_d \}$ be a discrete set of actions encoded as one-hot vectors.
The task is then to find:

\begin{equation}
\label{discrete-optim1}
\begin{aligned}
& \underset{a_1, ..., a_T}{\text{arg max}}
& & \sum_{t=1}^T f_R(f_S(\tilde{s}_{t-1}, a_{t-1}), a_t) \\
& \text{subject to}
& & a_t \in \{e_1,...,e_d \}
\end{aligned}
\end{equation}

Instead of requiring that the actions lie on the hypercube, we can relax the constraints and require that they lie on the simplex $\Delta^n$.
Note that the points on the simplex can be written as:

\begin{equation}
\begin{aligned}
\Delta^n &= \{ z : z_i \ge 0, \sum_i z_i = 1 \} \\
&= \{ z : z = \sigma(x), x \in \mathbb{R}^d \}
\end{aligned}
\end{equation}

where $\sigma$ represents the softmax function. The relaxed optimization problem can thus be reformulated as:

\begin{equation}
\label{bprop}
\begin{aligned}
& \underset{x_1, ..., x_T}{\text{arg max}}
& & \sum_{t=1}^T f_R(f_S(\tilde{s}_{t-1}, \sigma(x_{t-1})), \sigma(x_t)) \\
\end{aligned}
\end{equation}

A sequence of tokens can thus be chosen by minimizing the above loss function, quantizing each $\sigma(x_t)$ to the closest one-hot vector, and mapping back to the corresponding token.

This reformulation restricts the actions from the entire space to the simplex, however optimizing the actions by gradient descent may still yield points which are in the interior of the simplex rather than its vertices which are the one-hot vectors.
Our second modification is to modify the loss surface by adding Gaussian noise to the action vectors during training.
We then optimize the following loss function when training the forward model:

\begin{equation*}
\begin{aligned}
& \tilde{\mathcal{L}}(\theta_S, \theta_R) =
& & \frac{1}{M} \sum_{t=1}^M \Big[ \mathbb{E}_{\epsilon \sim \mathcal{N}(0, \sigma^2)} \ell(s_t, a_t + \epsilon, r_t, s_{t+1}) \Big] \\
\end{aligned}
\end{equation*}

This has the effect of making the loss surface more convex around the action vectors.
To see why, observe that the modified per-sample loss

\begin{equation*}
\tilde{\ell}(s_t, a_t, r_t, s_{t+1}) = \mathbb{E}_{\epsilon \sim \mathcal{N}(0, \sigma^2)} \Big[ \ell(s_t, a_t + \epsilon, r_t, s_{t+1}) \Big] \\
\end{equation*}

can be rewritten as:

\begin{equation*}
\begin{aligned}
\int \frac{1}{\sqrt{2\pi}\sigma} e^{\frac{-\epsilon^2}{2\sigma^2}} \ell(s_t, a_t + \epsilon, r_t, s_{t+1})d\epsilon  \\
\end{aligned}
\end{equation*}

The probability mass is highest at $\epsilon = 0$, which corresponds to the one-hot vector encoding $a_t$, and becomes progressively lower as we move further from the one-hot vector. This means that the model is presented with more points close to the one-hot vector during training, which will cause it to lower the loss surface at those points around that vector during its weight update. Points further from the one-hot vector will be presented less often, which means that the loss surface will be lowered less and less as we get further away from the one-hot vector which encodes $a_t$.
This results in a smoothing of the loss surface around the one-hot vectors encoding each of the actions, forming attractors which are easy to find during optimization.

\begin{figure}[t]
\vskip 0.2in
\begin{center}
\subfigure{\includegraphics[width=0.45\columnwidth]{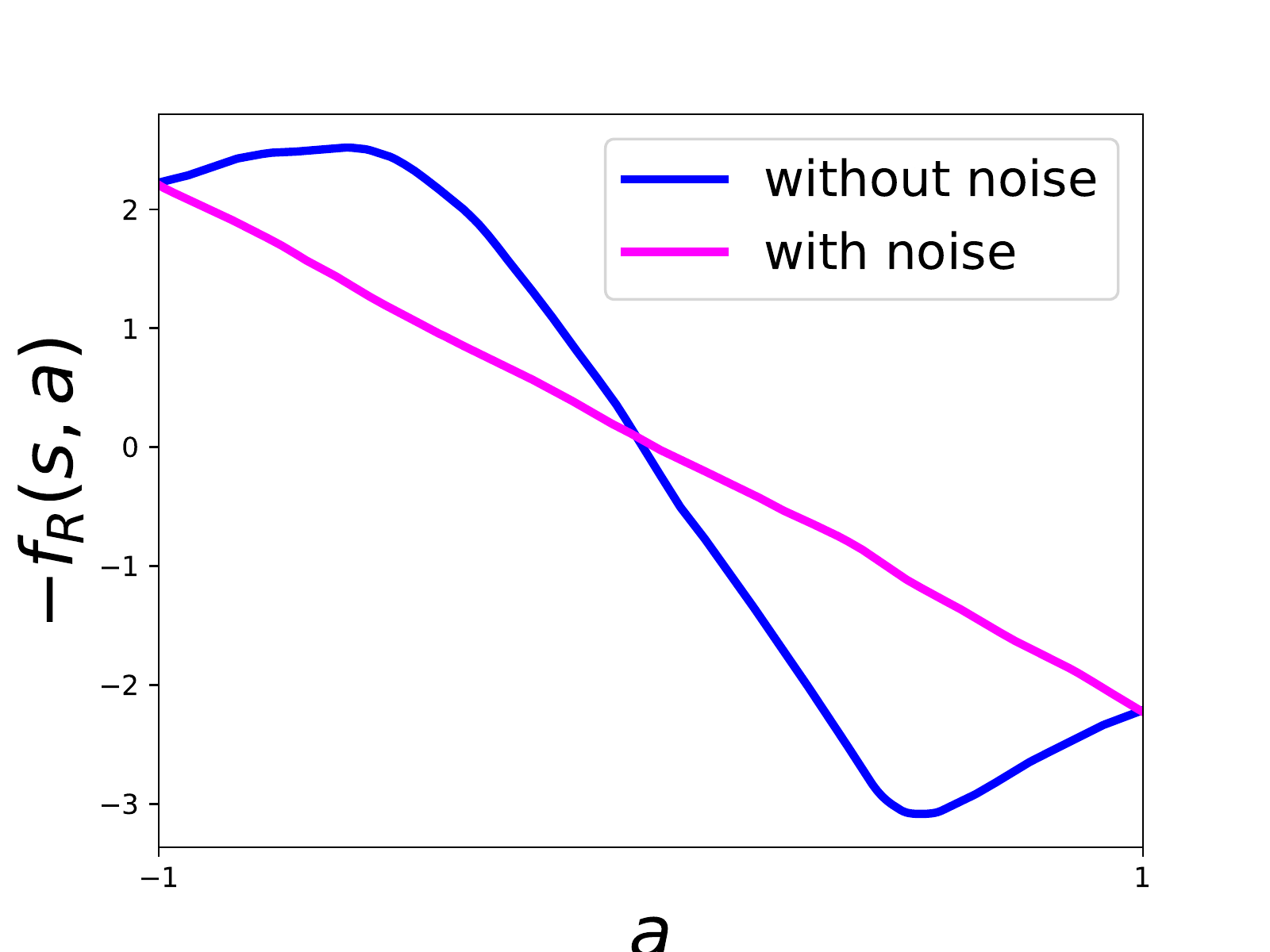}}
\subfigure{\includegraphics[width=0.45\columnwidth]{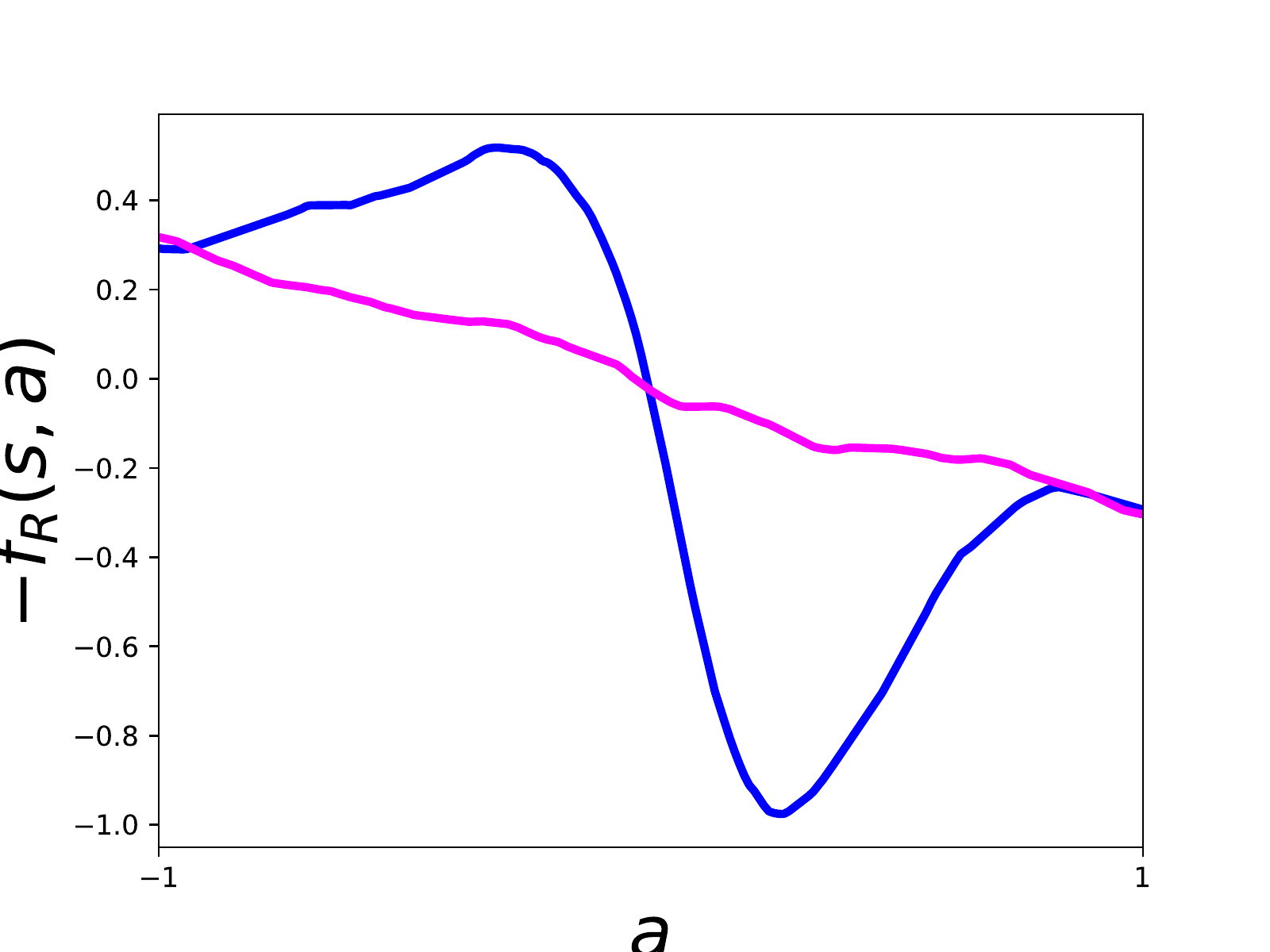}}
\subfigure{\includegraphics[width=0.45\columnwidth]{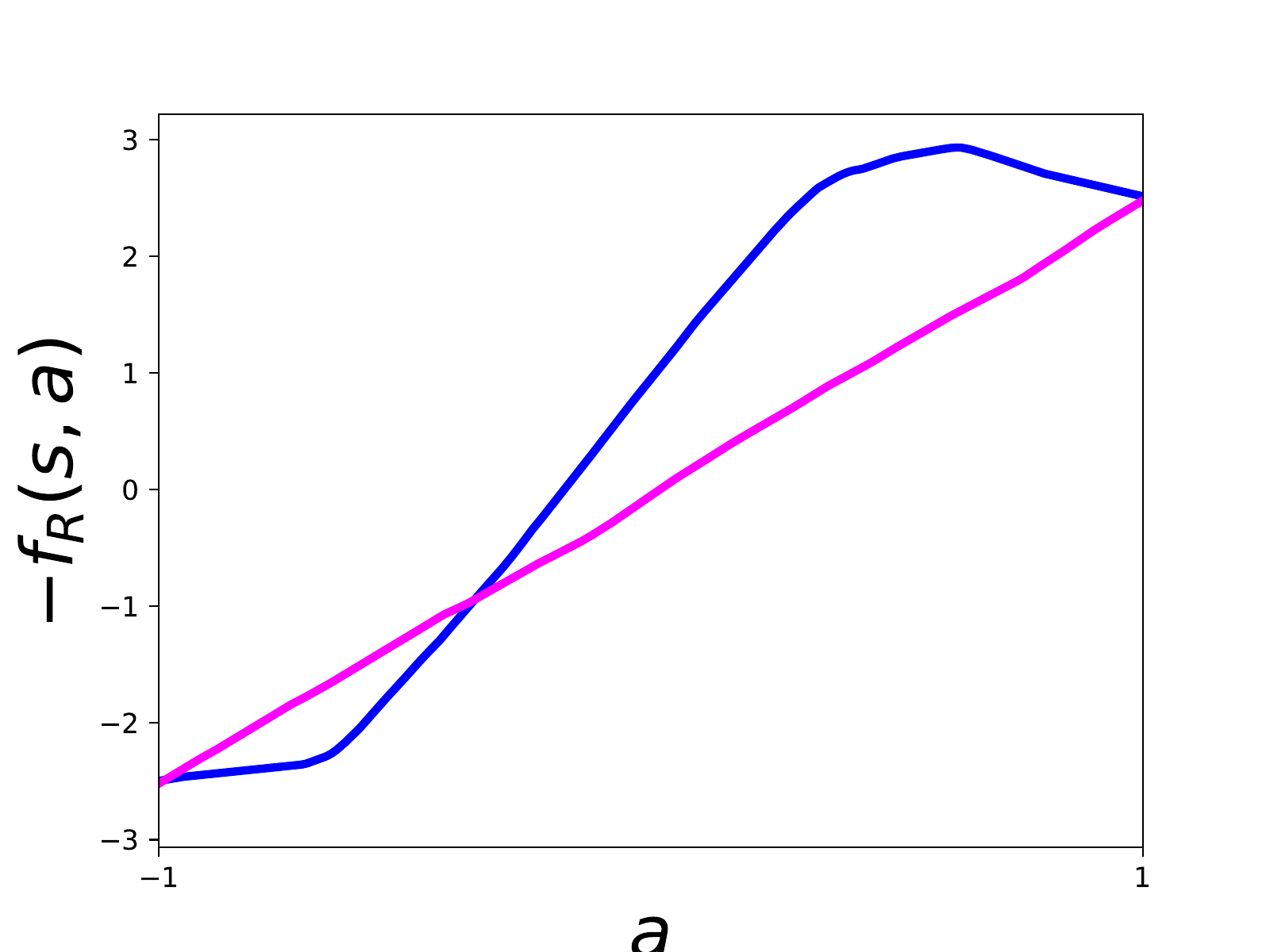}}
\subfigure{\includegraphics[width=0.45\columnwidth]{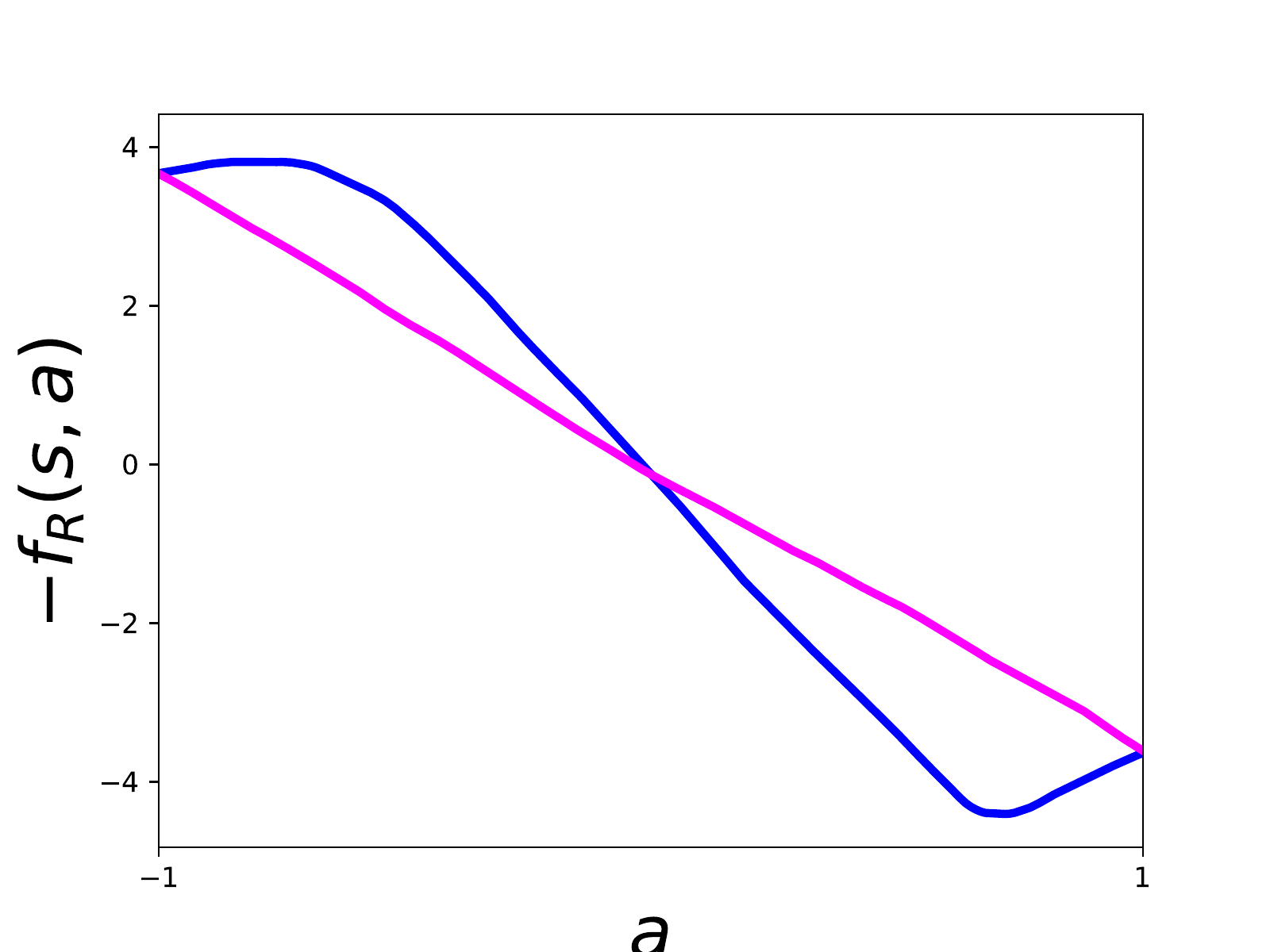}}
\caption{Predicted cost vs. action values for networks trained with and without input noise. Training with input noise smoothes the loss surface between the suboptimal and the optimal action.}
\label{toy-example}
\end{center}
\vskip -0.2in
\end{figure}

This method for inferring actions, which we here refer to as the Gradient-Based Planner, is detailed in Algorithm \ref{alg:algGBP}. In all our experiments, we set the variance of the added noise to $\sigma=0.25$.
To illustrate this effect we provide visualizations using a simple example.
Consider the simple one-step MDP given by:

\begin{equation}
r(s, a) =
\begin{cases}
      +s^3 & \mbox{ if } a = +1 \\
      -s^3 & \mbox{ if } a = -1 \\
   \end{cases}
\end{equation}

where $s \sim \mathcal{U}[-2, 2]$. The optimal action will be $a=+1$ if $s > 0$ and $a=-1$ if $s < 0$, and both actions become increasingly equivalent as $s$ gets closer to zero.
We trained two networks $f_R(s, a)$ to predict the reward from states and actions, using data where $s \sim \mathcal{U}[-2, 2]$, $a$ was $-1$ or $1$ with equal probability and $r(s,a)$ defined as above. One network was trained with Gaussian noise added to the input $a$ and the other was trained normally.
Both networks were 2-layer MLPs with 100 hidden units and ReLU activations and were trained with Adam \citep{ADAM}.
In Figure \ref{toy-example} we plot the predicted cost (negative reward) for different values of $a$ for both networks.
Note that both networks produce a similar prediction at the action values ${-1, 1}$ which are seen during training.
However, the cost surface between the action values $-1$ and $1$ for the network trained without noise exhibits local minima and maxima, whereas the predicted cost surface of the network trained with input noise is smooth and monotonically decreases from the suboptimal action to the optimal one.
This suggests that for a fixed state $s$, if we sample a random value of $a$ and follow the gradient along the cost surface, the result will be more likely to be the optimal action in the case of the network trained with noise than the network trained without noise.


\section{Self-Teaching a Policy Network}

Gradient-based planning requires solving an iterative optimization procedure every time an action or sequence of actions is required, which may be too slow for certain applications.
It may therefore be desirable to train a policy network which can quickly map states to actions at test time while still requiring few interactions with the environment.

We propose to do this by self-teaching a policy network using the learned model together with the dataset that the environment model was trained on.
This dataset consists of states, actions and rewards observed or experienced by the agent.
We can collect the states from this dataset and then use the learned environment model to infer what optimal actions would have been for each of these states, together with how states would evolve in response to these actions.
This creates a new dataset of state-action pairs which can be used to train a policy network $\pi(s)$ mapping states to actions inferred by the learned model in a supervised fashion. Note that this does not require any additional interactions with the environment as we only use the dataset used to train the forward model. We refer to this approach, shown in Algorithm \ref{alg:distGBP}, as DistGBP for Distilled Gradient-Based Planner.
We also note that such a policy network could be used to produce the initial rollouts in Algorithm \ref{alg:algGBP}, which would then be further refined by gradient descent.

\begin{algorithm}[tb]
   \caption{\modelname($s_0$)}
   \label{alg:algGBP}
\begin{algorithmic}
   \STATE {\bfseries Input:} Current state $s_0$, trained models $f_S, f_R$.
   \STATE {\bfseries Input:} Number of rollouts $K$, timesteps to unroll $T$, gradient steps $N$.
   \STATE Set $\tilde{s}_0 \leftarrow s_0$.
   \FOR{$k=1$ {\bfseries to} $K$}
   \STATE Sample $x_1, ..., x_T \sim \mathcal{N}(0, I)$
   \FOR{$i=1$ {\bfseries to} $N$}
   \FOR{$t=1$ {\bfseries to} $T$}
   \STATE $\tilde{a}_t \leftarrow \sigma(x_t)$
   \STATE $\tilde{s}_t \leftarrow f_S(\tilde{s}_{t-1}), \sigma(x_t))$
   \STATE $\tilde{r}_t \leftarrow f_R(\tilde{s}_{t-1}), \sigma(x_t))$
   \ENDFOR
   \STATE $\tilde{R} = \sum_t \tilde{r}_t$
   \FOR{$t=T$ {\bfseries to} $1$}
   \STATE $x_t \leftarrow x_t - \eta \frac{\partial \mathcal{\tilde{R}}}{\partial x_t}$
   \STATE $\tilde{a}_t \leftarrow \sigma(x_t)$
   \ENDFOR
   \ENDFOR
   \STATE $A_k \leftarrow (\tilde{a}_1, ..., \tilde{a}_T)$
   \STATE $S_k \leftarrow (s_0, \tilde{s}_1, ..., \tilde{s}_T)$
   \STATE $R_k \leftarrow \tilde{R}$.
   \ENDFOR
   \STATE {\bfseries Return:} Action sequence $A_k$ and state trajectory $S_k$ for which $R_k$ is largest.

\end{algorithmic}
\end{algorithm}

\begin{algorithm}[tb]
   \caption{Train \modelnamep}
   \label{alg:distGBP}
\begin{algorithmic}
   \STATE {\bfseries Input:} Dataset $\mathcal{E} = \{(s_t, a_t, r_t)\}_{t=1, ..., M}$ used to train models $f_S, f_R$.
   \STATE Build demonstration dataset: $\mathcal{D} \leftarrow \varnothing $
   \FOR{$i=1$ {\bfseries to} $N$}
   \STATE Sample $(s_t, a_t, r_t, s_{t+1}) \sim \mathcal{E}$
   \STATE $(s_t, ..., s_{t+T}), (a_t, ..., a_{t+T}) \leftarrow $ \modelname($s_t$)
   \FOR{$j=0$ {\bfseries to} $T$}
   \STATE $\mathcal{D} \leftarrow \mathcal{D} \cup (s_{t+j}, a_{t+j})$
   \ENDFOR
   \ENDFOR
   \STATE Train policy model:
   \REPEAT
   \STATE Sample $(s, a) \sim \mathcal{D}$
   \STATE Compute policy loss $\mathcal{L}_\pi(a, \pi(s))$
   \STATE Update $\pi \leftarrow \pi - \eta \nabla \mathcal{L}_\pi$.
   \UNTIL{converged}

\end{algorithmic}
\end{algorithm}

\section{Related Work}

The idea of planning a sequence of continuous actions by backpropagating along a policy trajectory has existed since the 1960's \citep{Kelley1960, Dreyfus1962}.
These methods were applied to settings where the state transition dynamics of the environment were known analytically and backward derivatives could be computed exactly, such as planning flight paths. Later works \citep{Schmidhuber1990, Jordan1992} explored the idea of backpropagating through learned, approximate forward models of the environment to plan actions for tasks such as parking a vehicle \citep{Nguyen1990}.

In recent years there have been several works revisiting model-based planning in the context of modern neural networks, for example vehicle navigation \citep{Hamrick2017} or robotics \citep{todorov2005generalized, abbeel2007application, todorov2012mujoco, kumar2016optimal}.
The work of \citep{I2A} used a learned model of the environment to plan sequences of discrete actions using imagined rollouts performed by a separate policy network, which are then encoded and fed as additional context to a model-free policy network. Our approach also uses discrete actions and a learned environment model, but differs in that we do not use reinforcement learning and instead correct initial random rollouts through gradient descent, which can then either be executed or used as training trajectories for a policy network.


There has been recent work in continuous relaxations of discrete random variables \citep{ConcreteDist, GumbelSoftmax}, that also uses a softmax to form a continuous approximation to a discrete set. These methods use a temperature parameter to anneal from the continuous formulation to the discrete one, which has the effect of pushing solutions towards the vertices of the simplex. In our approach we change the shape of the loss function of the forward model during training to have attractors at the vertices, rather than using a regularization term when optimizing the actions.


Our approach to training a policy network is related to the Dyna architecture introduced in \citep{Dyna}, which also uses a learned model of the environment to perform policy updates without needing to interact with the environment. This was introduced in the context of $Q$-learning in the tabular setting.
Our policy updates are related to imitation learning \citep{Pomerleau91} where an agent is trained to imitate trajectories provided by an expert; here the training trajectories are provided by our gradient-based planning algorithm.
The work of \citep{AtariMCTS} used Monte-Carlo Tree Search (MCTS) together with an Atari simulator to generate high-quality action sequences for different games which were then used to train an agent through imitation learning. This is related to our approach which also uses a slower planning method to generate trajectories offline which are then used to train a fast policy network.
However, we use a learned model of the environment rather than a ground-truth simulator to generate trajectories, and our method of inferring action sequences is different.
Also related are policy distillation methods where a complex neural network is approximated by a simpler and typically faster one \citep{Hinton15, PolicyDistillation}. Here, we train a network to approximate the results of a planning procedure rather than the output of another network.

\section{Experiments}

We tested our approach on two domains: one with a purely discrete action space, and one where the actions space contains both discrete and continuous actions.
In these experiments, we evaluate different methods according to three measures: how well the method solves the task, the number of interactions with the environment needed to achieve good performance, and the speed at inference time.
For each experiment, we evaluate both the gradient-based planner described in Algorithm 2, which we denote \modelname, and the policy network trained to imitate trajectories from this model as described in Section 4, which we call \modelnamep.

\begin{figure}[t]
\vskip 0.2in
\begin{center}
\subfigure{\includegraphics[width=0.3\columnwidth]{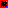}}
\subfigure{\includegraphics[width=0.3\columnwidth]{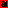}}
\subfigure{\includegraphics[width=0.3\columnwidth]{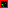}}
\subfigure{\includegraphics[width=0.3\columnwidth]{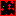}}
\subfigure{\includegraphics[width=0.3\columnwidth]{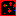}}
\subfigure{\includegraphics[width=0.3\columnwidth]{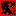}}
\caption{Examples of $8\times8$ and $16\times16$ Gridworld maps.}
\label{gridworld-pics}
\end{center}
\vskip -0.2in
\end{figure}

\subsection{Gridworld Domains}

The first task we evaluated our approach on was the Gridworld domain introduced in \citep{VIN}.
In these tasks, an agent is placed in a map and must make its way to a goal while avoiding obstacles.
The MDP is structured as follows: states are represented by $3$-channel images of size $8 \times 8$ or $16 \times 16$ where the first channel represents obstacle locations, the second the goal location and the third the agent's current location (examples are shown in Figure \ref{gridworld-pics}). The agent can perform 4 actions: move $\{\texttt{north}, \texttt{south}, \texttt{east},\texttt{west}\}$ and receives a reward of $+1$ for reaching the goal, a reward of $-1$ for hitting an obstacle (after which the episode ends), and a reward of $-0.01$ for every other timestep to encourage it to reach the goal quickly. The test set consists of maps which are different from those used for training.

For each gridworld size we trained an action-conditional forward model using 10K episodes where the agent followed a uniform random policy over actions.
The architecture consisted of a convolutional encoder whose output was combined with a learned embedding of the action vector, which was then fed to a convolutional decoder to predict the next state, as well as a second convolutional network followed by a fully-connected layer with scalar output and hyperbolic tangent to predict the reward. We used 16 feature maps for all convolutional layers and 16 hidden units for the fully-connected layers. All networks were trained with Adam \citep{ADAM} using a learning rate of 0.001.

As a first experiment, we compared \modelname to Monte-Carlo Tree Search (MCTS) \citep{MCTS}. MCTS is a discrete planning algorithm where a search tree is selectively expanded in directions which are likely to be most promising, as determined by simulated playouts from the leaves of the tree. This method can be shown to achieve optimal performance given a perfect environment model and enough simulated rollouts, and has been successfully used in contexts such as games \citep{AlphaGo}, combinatorial optimization \citep{CombOptUCT} and scheduling \citep{MCTSBus}. As a simulator, we used the same learned model of the environment as for \modelname. Both methods can trade accuracy for computation time by performing larger numbers of rollouts.

\begin{figure}[t]
\vskip 0.2in
\begin{center}
\subfigure{\includegraphics[width=0.48\columnwidth]{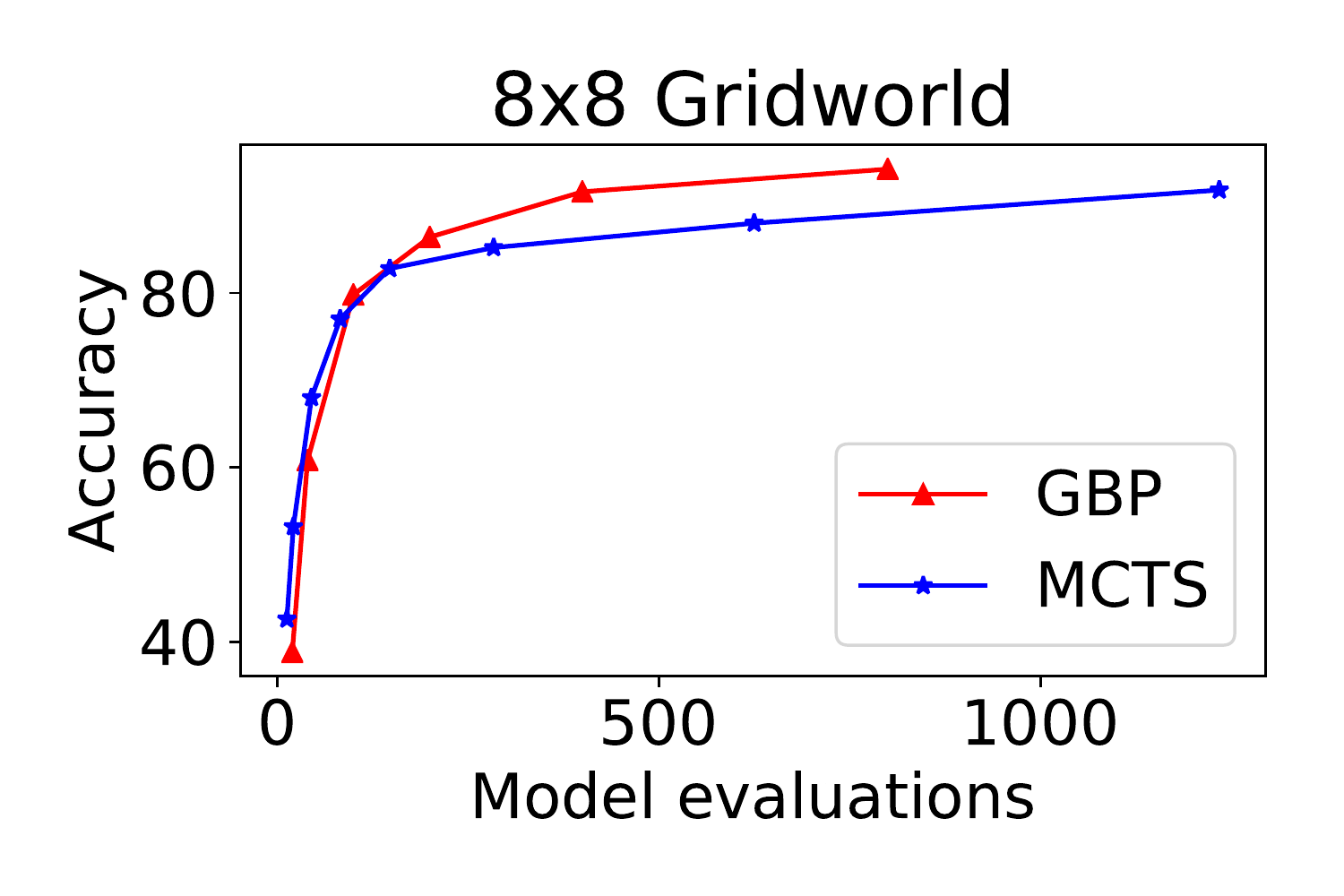}}
\subfigure{\includegraphics[width=0.48\columnwidth]{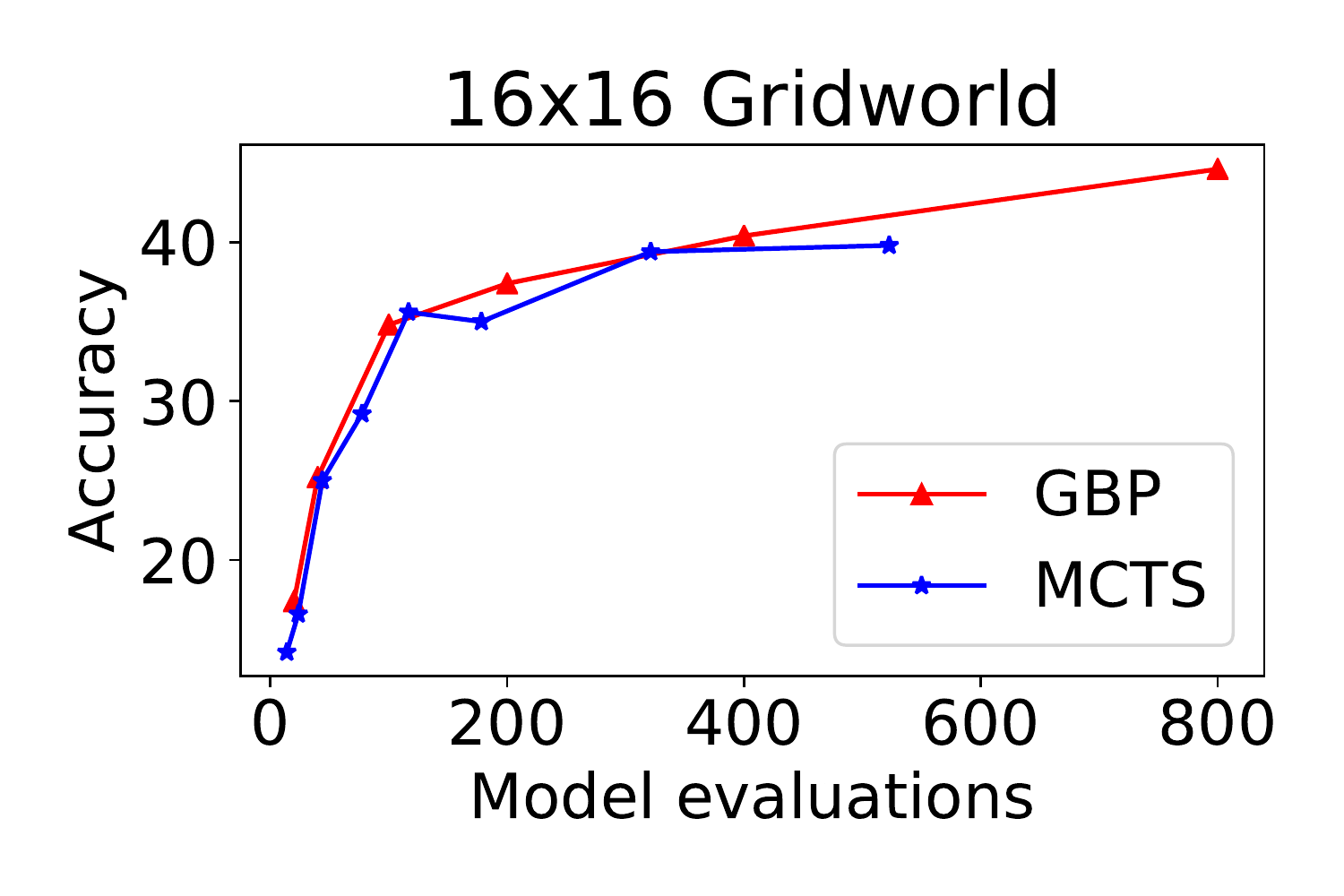}}
\caption{Accuracy vs. computation for \modelname and MCTS, measured in number of passes through the forward model.}
\label{computation}
\end{center}
\vskip -0.2in
\end{figure}

Figure \ref{computation} shows the tradeoff between accuracy and computation time for \modelname and MCTS, measured in the number of forward and backward passes through the environment model.
For MCTS we computed performance using $\{10, 20, 50, 100, 200, 400\}$ rollouts, for \modelname we used 10 gradient steps and $\{1, 2, 5, 10, 20, 40\}$ rollouts.
All results are averaged over 500 trials.
We see that both methods require a similar number of passes through the model to achieve a given level of accuracy, which suggests that \modelname can discover sequences of discrete actions through gradient descent of similar quality to those discovered by a strong discrete planning algorithm.
We also note that \modelname can be easily parallelized on a GPU by treating different rollouts as samples in a minibatch, which allows us to increase the accuracy at little computational cost, whereas MCTS is more challenging to parallelize due to the sequential nature of the updates to the tree policy and the variable length of simulated playouts \citep{ParallelUCT, SurveyMCTS}.

We next compared different approaches in terms of best accuracy, planning time and number of environment interactions, shown in Table \ref{gridworld-results}.
We report the TRPO results of \citep{VIN}, which use a CNN model trained with Trust Region Policy Optimization \citep{TRPO} together with a curriculum whereby easier maps (with the goal placed close to the agent) are shown early in training and harder maps are shown later.
For the $8 \times 8$ maps we also include results using the OpenAI Baselines \citep{OpenAIBaselines} implementation of TRPO using the same architecture since the published results do not include the number of training steps.
To train the \modelnamep, we generated 24K trajectories using \modelname for the $8 \times 8$ maps and 75K trajectories for the $16 \times 16$ maps, and trained a policy model with supervised learning. The policy model for $8 \times 8$ maps was a 2-layer CNN with 16 feature maps, followed by a fully-connected layer with 16 hidden units; for the $16 \times 16$ maps we used a 3-layer CNN with 64 feature maps and 64 hidden units in the fully-connected layer.

\begin{table}[t]
\caption{Performance on Gridworld Tasks. Time indicates the average number of passes through a policy or forward model required to choose an action at inference time. Env. Steps indicates the number of interactions with the environment to train the model. TRPO* indicates results from \citep{VIN}. All other results are averaged over 500 trials.}
\label{gridworld-results}
\vskip 0.15in
\begin{center}
\begin{small}
\begin{sc}
\resizebox{\columnwidth}{!}{
\begin{tabular}{l|lllll}
\toprule
Map & Method & Acc. & Time (s) & Env. Steps \\
\midrule
\multirow{6}{*}{$8 \times 8$}
 & TRPO*    & 86.9 & $<0.001$ & - \\
 & TRPO (ours)    & 82.4 & $< 0.001$ & 22M \\
 & \modelname  & 94.0 & 0.03 & 54K \\
 & \modelnamep    & 91.4 & $< 0.001$ & 54K \\
 & \modelname (no noise) & 25.6 & 0.03 & 54K \\
\hline
\hline
\multirow{6}{*}{$16 \times 16$}
 & TRPO*    & 33.1 & $< 0.001$ & 3M \\
 & \modelname     & 66.4 & 0.51 & 110K \\
 & \modelnamep     & 52.6 & $< 0.001$ & 110K        \\
 & \modelname (no noise)    & 07.8 & 0.51 & 110K \\
\bottomrule
\end{tabular}
}

\end{sc}
\end{small}
\end{center}
\vskip -0.1in
\end{table}

Both implementations of model-free TRPO achieve good performance on the $8 \times 8$ gridworld task and are able to perform fast inference \footnote{Since both models have the same architecture, we report the same inference time for the published results as for our model.}. We found that our TRPO model only achieved comparable performance to the published TRPO model after a large number of steps (22 million). The other methods use a learned environment model, which is here done using comparably fewer environment interactions.
This is sufficient to achieve good performance for \modelname given enough planning time.
Our self-taught policy network is furthermore able to achieve comparable performance while being as fast as the reactive policy at inference time and using few environment interactions. Furthermore, it does not require a curriculum as it is trained in a supervised manner on trajectories that are optimized over many timesteps and thus capture longer-term dependencies between actions and rewards.

The $16 \times 16$ gridworld task is more challenging as it requires finding longer paths.
Here we increased the unrolling depth of the forward model from 10 to 15 in order to capture longer-term dependencies and increased the number of rollouts to 1000.
\modelname here performs significantly better than the model-free method at the cost of additional computation time at inference.
\modelnamep is also able to perform significantly better than TRPO, while having the same speed and using fewer environment interactions.

\begin{figure}[t]
\vskip 0.2in
\begin{center}
\subfigure{\includegraphics[width=0.1\columnwidth]{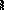}}
\hspace{0.01in}
\subfigure{\includegraphics[width=0.1\columnwidth]{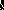}}
\hspace{0.01in}
\subfigure{\includegraphics[width=0.1\columnwidth]{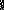}}
\hspace{0.01in}
\subfigure{\includegraphics[width=0.1\columnwidth]{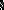}}
\hspace{0.01in}
\subfigure{\includegraphics[width=0.1\columnwidth]{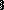}}
\hspace{0.01in}
\\
\subfigure{\includegraphics[width=0.1\columnwidth]{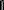}}
\hspace{0.01in}
\subfigure{\includegraphics[width=0.1\columnwidth]{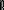}}
\hspace{0.01in}
\subfigure{\includegraphics[width=0.1\columnwidth]{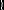}}
\hspace{0.01in}
\subfigure{\includegraphics[width=0.1\columnwidth]{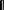}}
\hspace{0.01in}
\subfigure{\includegraphics[width=0.1\columnwidth]{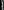}}
\hspace{0.01in}
\caption{Action sequences inferred by backprop using Algorithm 2 using a model trained with noise (top) and without noise (bottom). Each row represents an action vector in the sequence, columns represent the indices of discrete actions.}
\label{actionvecs}
\end{center}
\vskip -0.2in
\end{figure}

We also include results for forward models trained without noise. For both Gridworld datasets, using a model trained without noise causes a substantial decrease in performance.
Figure \ref{actionvecs} shows action vectors optimized with backprop for the same set of input states using a forward model trained with and without noise.
In the first case most of the inferred action vectors are close to the one-hot vectors, which are consistent with the input distribution the model was trained on.
In the second case, the solution does not resemble any input seen during training which causes the model to produce an incorrectly large reward estimate, in the same way as adversarial examples can cause a model to predict an incorrect image with high confidence.

\subsection{Spaceship Domain}

Having validated that our approach can indeed be used to optimize discrete actions by gradient descent, we next tested it on a domain where the action space contains both discrete and continuous actions.
In settings with both types of actions, tree search methods such as MCTS cannot be applied as the search space is effectively infinite and unmodified optimal control methods cannot be straightforwardly applied due to the presence of discrete actions.
We used a spaceship domain inspired by recent works \citep{Hamrick2017, Pascanu17}, where the agent must pilot a spaceship in the presence of planets and their gravitational forces.
In our task, the agent must pilot its ship to make contact with one of three different colored waypoints using its thrusters and emit a colored signal the same color as the waypoint. Applying thrust is a continuous action while emitting a colored signal is a discrete action, with one possibity for each color in addition to no signal. Furthermore, the agent must avoid getting too close to the planet's gravitational field which may cause it to crash into the planet.

The agent receives a reward of $+1$ for emitting the same color as a waypoint it touches, a reward of $-1$ emitting a different color, a reward of $-0.1$ for emitting a color elsewhere than at a waypoint, and a reward of $-1$ for coming into contact with a planet. At every time step, the agent's position and velocity are updated as a function of the thrust vector and the gravitational force applied by the planet using the Euler method (details are provided in the Appendix).
The agent must therefore learn to navigate to a waypoint while avoiding the planets (which requires optimizing continuous actions) and execute the correct discrete action once it gets there. Each episode is of length 80 time steps and the agent keeps receiving rewards (positive or negative) until the end of the episode.

\begin{figure}[t]
\vskip 0.2in
\begin{center}
\subfigure{\includegraphics[width=0.45\columnwidth]{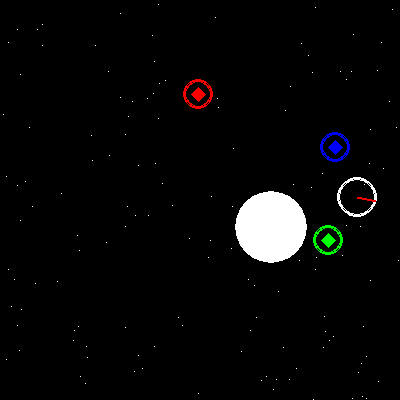}}
\subfigure{\includegraphics[width=0.45\columnwidth]{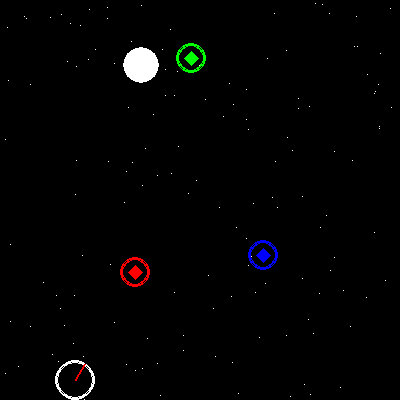}}
\caption{Two maps in the Spaceship environment. Solid white circle is the planet, colored circles with diamonds are waypoints, white outlined circle is the spaceship with a red line representing the thrust vector.}
\label{space-image}
\end{center}
\vskip -0.2in
\end{figure}

We compared three approaches: \modelname, \modelnamep,  and an Advantage Actor-Critic (A2C) agent, a state-of-the-art model-free method \citep{A3C}.
We trained a forward model on 10K episodes of the agent following a random policy where its continuous thrusters were fired according to an isotropic Gaussian distribution and its discrete actions followed a uniform categorical one (architecture and training details are in the Appendix).  The A2C agent consisted of a 4-layer actor network and a 4-layer critic network, each with 512 hidden units and ReLU activations. The actor network output two heads: a categorical distribution over discrete actions and a 2D diagonal gaussian distribution over continuous actions. We trained the A2C model with 16 parallel workers and optimized the learning rate over the range $\{0.0001, 0.001, 0.01 \}$ and the entropy regularization coefficient over $\{0, 0.01, 0.05\}$. The \modelnamep policy network had the same architecture as the actor network in the A2C model.

Average rewards for the policies learned by the different models as well as a random policy are shown in Table \ref{space-results}. Videos of the the \modelnamep agent at can be seen at \url{https://youtu.be/9Xh2TRQ_4nM} and the A2C agent at \url{https://youtu.be/XLdme0TTjiw}.
The A2C model is able to learn a policy where it moves away from the planet, which enables it to achieve a significant improvement in average reward over the baseline. However, it does not learn to navigate towards the waypoints or use its signals to collect positive rewards and only tries to minimize negative reward by avoiding the planet's gravity.
The \modelnamep agent significantly outperforms the A2C model in terms of average reward, and learns interesting behaviors such as moving away from the planet, moving towards the waypoints when it gets sufficiently close and turning the right color when it reaches them. This requires learning dependencies between continuous and discrete actions, since turning the wrong color when touching a waypoint incurs a negative reward. It is also able to accurately compensate for the gravitational pull of the planet when it touches a waypoint, applying the correct thrust vector depending on the planet size and location to remain stationary and thus maximize its reward.
This indicates that the policy network is able to leverage the environment dynamics learned by the forward model, which are themselves reflected in the state-action trajectories the policy network is trained on. Note that \modelname performs similarly to \modelnamep in terms of reward, but has considerably higher inference time.

\begin{table}[t]
\caption{Performance of different methods on the Spaceship Task. Reward is measured in average reward per episode.}
\label{space-results}
\vskip 0.15in
\begin{center}
\begin{small}
\begin{sc}
\begin{tabular}{lcccr}
\toprule
Method & Average Reward & Time (s) & Env. Steps \\
\midrule
Random   & -62.7 & - & 0 \\
A2C   & -19.2  & 0.01 & 3.8M \\
\modelname   & 11.1 & 0.19 & 800K \\
\modelnamep   & 12.2 & 0.01 & 800K \\
\bottomrule
\end{tabular}
\end{sc}
\end{small}
\end{center}
\end{table}


\section{Conclusion}

In this work, we have introduced a novel method for performing gradient-based planning in discrete action spaces and shown that it can effectively be used both in discrete action settings as well as settings which combine discrete and continuous actions where other methods are not easily applicable.
Furthermore, we have shown that the iterative procedure required to obtain high-quality action sequences through gradient descent is not an obstacle for real-time applications, as the planning policy can be approximated by a fast feedforward network trained to imitate optimal trajectories produced by the model. Taken together, these steps provide a general approach for deriving agents capable of executing sophisticated policies in real time which do not require large amounts of environment interaction.


\nocite{langley00}

\bibliography{icml2018}
\bibliographystyle{icml2018}

\appendix

\section{Appendix}

\subsection{Training details for Gridworld tasks}

\subsection{Spaceship Environment}

The state representation at each timestep consists of the concatenation of the following vectors:

\begin{equation*}
s = [x_s, v_s, r_s, x_{w_1}, r_{w_1}, x_{w_2}, r_{w_2} x_{w_3}, r_{w_3}, x_p, r_p]
\end{equation*}

where $x_s, x_p, x_{w_1}, x_{w_2}, x_{w_3}$ are the 2D position vectors of the spaceship, planet and waypoints respectively; $v_s$ is the velocity of the spaceship, and $r_s, r_p, r_{w_1}, r_{w_2}, r_{w_3}$ are the radii of the spaceship, planet and waypoints.
At each timestep, the force of gravity applied to the spaceship is computed as:

\begin{equation*}
F = G \frac{m_p m_s}{r^2}(x_p - x_s)
\end{equation*}

where $G=0.015$, $m_s = 1$ and $m_p = 20r_p$. Acceleration is then computed as

\begin{align*}
a_s = \frac{F_p - dv_x + u}{m_s} \\
\end{align*}

where $d=0.1$ is a damping constant and $u$ is the 2D thrust vector (a continuous action given by the agent).
Position and velocity are then updated using a simulation step size of $\epsilon = 4$:

\begin{align*}
x_s \leftarrow x_s + \epsilon v_s \\
v_s \leftarrow v_s + \epsilon a_s
\end{align*}

\subsection{Training details for spaceship task}

The forward model consisted of a state encoder, a state predictor and a reward predictor, which were all 2-layer MLPs with 512 hidden units and PReLU activations, as well as a linear action encoder which was added to the state encoding. The model was trained using Adam with learning rate 0.0001. We unrolled the forward model for 40 timesteps, providing it only the initial state in addition to the action sequence so that it used its predictions as subsequent state inputs as shown in Figure \ref{fwd-train}. This makes the learning problem much more challenging and helps encourage the model to be robust to its previous prediction errors.
During inference, \modelname performed 20 gradient steps with 20 rollouts. These same hyperparameters were used to generate the trajectories \modelnamep was trained on.

As a policy network for \modelnamep, we used a 4-layer MLP with 512 hidden units and ReLU activations with two heads: a softmax over actions and a linear layer mapping to 2D continuous actions. This is the same network architecture as the actor network in the A2C model.

\end{document}